%% file: main.tex
\documentclass{article}
\PassOptionsToPackage{numbers,sort&compress}{natbib}
\usepackage[preprint]{neurips_2024}
\usepackage[T1]{fontenc}
\usepackage[utf8]{inputenc}
\usepackage{times,latexsym,microtype,graphicx}
\usepackage{amsmath,amssymb,booktabs,tabularx,array,enumitem,xcolor,caption}
\usepackage[hidelinks]{hyperref}
\graphicspath{{figures/}{assets/}}
\hypersetup{pdftitle={Vibe-Trading: Towards Agent-Native Trading},pdfauthor={Haozhe Wu, Zhonghang Li and Chao Huang}}
\setlist{leftmargin=1.4em,itemsep=2pt,topsep=3pt,parsep=0pt}
\newcolumntype{Y}{>{\raggedright\arraybackslash}X}
\usepackage[ruled]{algorithm}
\usepackage[noend]{algpseudocode}
\usepackage{colortbl}%
\usepackage{pgfplots}
\pgfplotsset{compat=1.18}
\usepgfplotslibrary{groupplots}
\usetikzlibrary{positioning,calc,shapes.geometric,shapes.misc,arrows.meta}
\definecolor{vtblue}{HTML}{2A78D6}
\definecolor{vtorange}{HTML}{EB6834}
\definecolor{vtaqua}{HTML}{1BAF7A}
\definecolor{vtblueL}{HTML}{86B6EF}
\definecolor{vtblueD}{HTML}{1C5CAB}
\definecolor{vtgood}{HTML}{0CA30C}
\definecolor{vtwarn}{HTML}{FAB219}
\definecolor{vtcrit}{HTML}{D03B3B}
\definecolor{vtink}{HTML}{0B0B0B}
\definecolor{vtink2}{HTML}{52514E}
\definecolor{vtmuted}{HTML}{898781}
\definecolor{vtgrid}{HTML}{E1E0D9}
\definecolor{vtaxis}{HTML}{C3C2B7}
\definecolor{vtsurf}{HTML}{F6F5F2}
\tikzset{subcap/.style={font=\footnotesize, text=vtink, align=center, inner sep=1pt}}%
\pgfplotsset{vtaxis/.style={axis line style={vtaxis}, tick style={vtaxis}, tick label style={font=\scriptsize, text=vtink2}, label style={font=\scriptsize, text=vtink2}, title style={font=\scriptsize\bfseries, text=vtink}, major grid style={vtgrid, line width=0.3pt}, legend style={font=\scriptsize, draw=none, fill=none, text=vtink2}}}
\newcommand{\system}{Vibe-Trading}

\makeatletter
\renewcommand{\@noticestring}{Technical Report}
\makeatother
\title{Vibe-Trading: Towards Agent-Native Trading}
\author{Haozhe Wu \quad Zhonghang Li \quad Chao Huang\thanks{Corresponding author.}\\
The University of Hong Kong\\
\texttt{\{wu030618, bjdwh.zzh, chaohuang75\}@gmail.com}\\
\textcolor{blue}{\url{https://github.com/HKUDS/Vibe-Trading}}}
\begin{document}
\raggedbottom
\maketitle
\begin{abstract}
Existing AI agents can retrieve information, call tools, run experiments, and produce multi-step outputs. Yet most are built around relatively static workflows and struggle to preserve reliable state as the environment changes. Financial trading makes this limitation especially consequential: prices, news, market conditions, portfolio positions, and available capital evolve continuously. A result is useful only when it remains tied to the correct time, data snapshot, and execution context. We present \textbf{Vibe-Trading}, a comprehensive agent-native trading framework and all-in-one workspace for building, evaluating, and evolving agents for financial trading. It unifies market research, financial data, strategy generation, backtesting, portfolio analysis, paper trading, multi-agent collaboration, and optional broker execution. Its central contribution is an architecture that keeps results, evidence, research history, and trading authority connected throughout the research-to-action process. Each result retains its source, temporal scope, configuration, and producing run. Shared evidence grounding enables verification, versioned research records support reproducible continuation, and an explicit authority gate separates strategy research from live execution. Together, these designs provide a practical foundation for trading agents whose results can be traced, reproduced, and acted on only under explicit authority.
\end{abstract}

\input{sections/01_introduction}

\input{sections/02_preliminaries}
\input{sections/02_framework}

\input{sections/03_execution_evaluation}

\input{sections/04_insights}

\input{sections/05_related_work}
\input{sections/05_conclusion}
\clearpage
\begingroup
\small
\setlength{\bibsep}{0pt plus 0.5pt}
\enlargethispage{\baselineskip}
\bibliographystyle{plainnat}
\bibliography{references}
\endgroup
\clearpage
\appendix
\input{sections/appendix_evidence}

\end{document}

%% file: sections/01_introduction.tex
\section{Introduction}
\label{sec:introduction}

AI agents are capable of retrieving information, calling tools, running experiments, and completing multi-step tasks. However, many existing agent frameworks are designed around fixed task flows and limited state tracking. They may work well when the environment and task assumptions remain stable, but become less reliable when data, observations, system states, or available actions change over time. This raises a question: how can an agent framework maintain provenance and state consistency across tool calls, intermediate results, evolving research states, and downstream actions, while keeping each decision grounded in the right evidence and authorization context?

Financial trading makes this challenge especially difficult. Prices, news, market conditions, portfolio positions, and available capital can change continuously. A result that is correct at one moment may become misleading when its time period, data snapshot, portfolio state, or producing run is unclear. Reliable trading agents must therefore preserve more than numerical outputs. They must also retain the context that makes those outputs interpretable and reusable.

Trading also requires agents to continuously improve and evolve their strategies. They must decide which signals to trust, which hypotheses to test, when to abandon a weak approach, and how to adapt to changing market conditions. These decisions are costly: every experiment consumes time, tokens, and computation, while a failed strategy may introduce real financial or operational risk. A capable executor is not enough if its policy repeatedly sends it in the wrong direction. This motivates an architecture that connects strategy learning, financial computation, evidence, research history, and authorized actions throughout the research-to-action process.

Recent work has made important progress toward tool-using and financial agents. Tool-using agents support flexible planning and execution~\citep{yao2023react}, while financial agents incorporate domain knowledge, memory, and specialist collaboration~\citep{yang2024finrobot,yu2023finmem,xiao2024tradingagents}. Other systems study evidence attribution and how generated answers retain supporting sources~\citep{gao2023alce,gao2023rarr}, while financial environments provide standardized interfaces for market data and brokerage interaction~\citep{luo2026qfinzero}. However, these capabilities are usually distributed across separate tools or modules. They do not yet provide a unified environment for building, evaluating, self-improving, and operating trading agents.

We present \textbf{Vibe-Trading}, an agent-native trading framework and all-in-one workspace for building, evaluating, and evolving trading agents. By \emph{agent-native} we mean financial results, their supporting evidence, research state, and action permissions are explicit objects agents can inspect and use. It integrates market research, financial data, strategy generation, backtesting, portfolio analysis, paper trading, multi-agent collaboration, and optional broker execution. Its architecture connects these capabilities through shared representations of results, evidence, research history, portfolio state, and trading authority. Each result retains its data range, configuration, portfolio snapshot, and producing run, allowing agents to compare strategies and continue research from a well-defined state.

Vibe-Trading further introduces an evidence-grounded execution design. Stored artifacts, model-visible evidence, and checker-admitted support are represented separately but linked through explicit references to the producing tool, engine, run, and artifact. This prevents value-only matching from validating claims from the wrong run. A separate authority gate keeps strategy research independent from broker execution and checks actions against account permissions and constraints. Together, these mechanisms provide a unified foundation for reliable trading-agent research and continuous strategy improvement in dynamic financial environments.

Our \textbf{Vibe-Trading} is organized around three components. $\bullet$ \textbf{Identity-Preserving Evidence Grounding}: links each result to its source, context, configuration, and producing run. $\bullet$ \textbf{Unified Research-to-Action Architecture}: carries these identified results through strategy evaluation, portfolio analysis, research continuation, and authorized execution. $\bullet$ \textbf{Reproducible and Auditable System Implementation}: exposes the mechanisms through inspectable interfaces, reproducible tests, and source-level evidence. Together, these components form a coherent foundation that allows trading agents to preserve meaning as they research, improve strategies, and move toward action. They are realized by two layers: a financial research \emph{workspace} that computes and preserves identified results, and an agent \emph{harness} that controls how a policy uses them (Section~\ref{sec:framework}).

%% file: sections/02_preliminaries.tex
\section{Preliminaries}\label{sec:preliminaries}
We consider financial research tasks, such as option pricing, portfolio risk analysis, and strategy comparison. An LLM trading agent addresses a task $q$ through sequential tool interactions~\citep{yao2023react,schick2023toolformer}:
\[
a_t \sim \pi\bigl(\cdot \mid \mathcal{H}(q,\, a_{<t},\, o_{<t})\bigr),\qquad
(o_t,\, s_{t+1}) = \mathcal{W}\bigl(s_t,\, a_t\bigr).
\]
At step $t$, the harness $\mathcal{H}$ constructs the context from the task and prior interactions, and the policy $\pi$ selects an action $a_t$, such as retrieving data, performing a calculation, saving a result, or proposing an answer. The workspace $\mathcal{W}$ provides financial data, computation services, and persistent research records. It executes $a_t$ on state $s_t$, returning an observation $o_t$---a result or an error---and the updated state $s_{t+1}$. A task is complete when the requested results are saved and reported correctly.

LLMs offer general-purpose reasoning and planning capabilities across diverse tasks, but their effectiveness depends on the information and operations available through the surrounding environment. In financial research, a key practical bottleneck is access to relevant observations, computational tools, and prior results, together with the information needed to interpret and use them correctly. Financial meaning, temporal scope, and assumptions become particularly important as prices, positions, and research conditions evolve. We therefore keep $\pi$ replaceable and design $\mathcal{W}$ and $\mathcal{H}$ to expose, preserve, and organize these resources, supporting informed decisions grounded in traceable evidence.

%% file: sections/02_framework.tex
\section{The Vibe-Trading Framework}\label{sec:framework}
\begin{figure}[!ht]
\centering
\includegraphics[width=\textwidth]{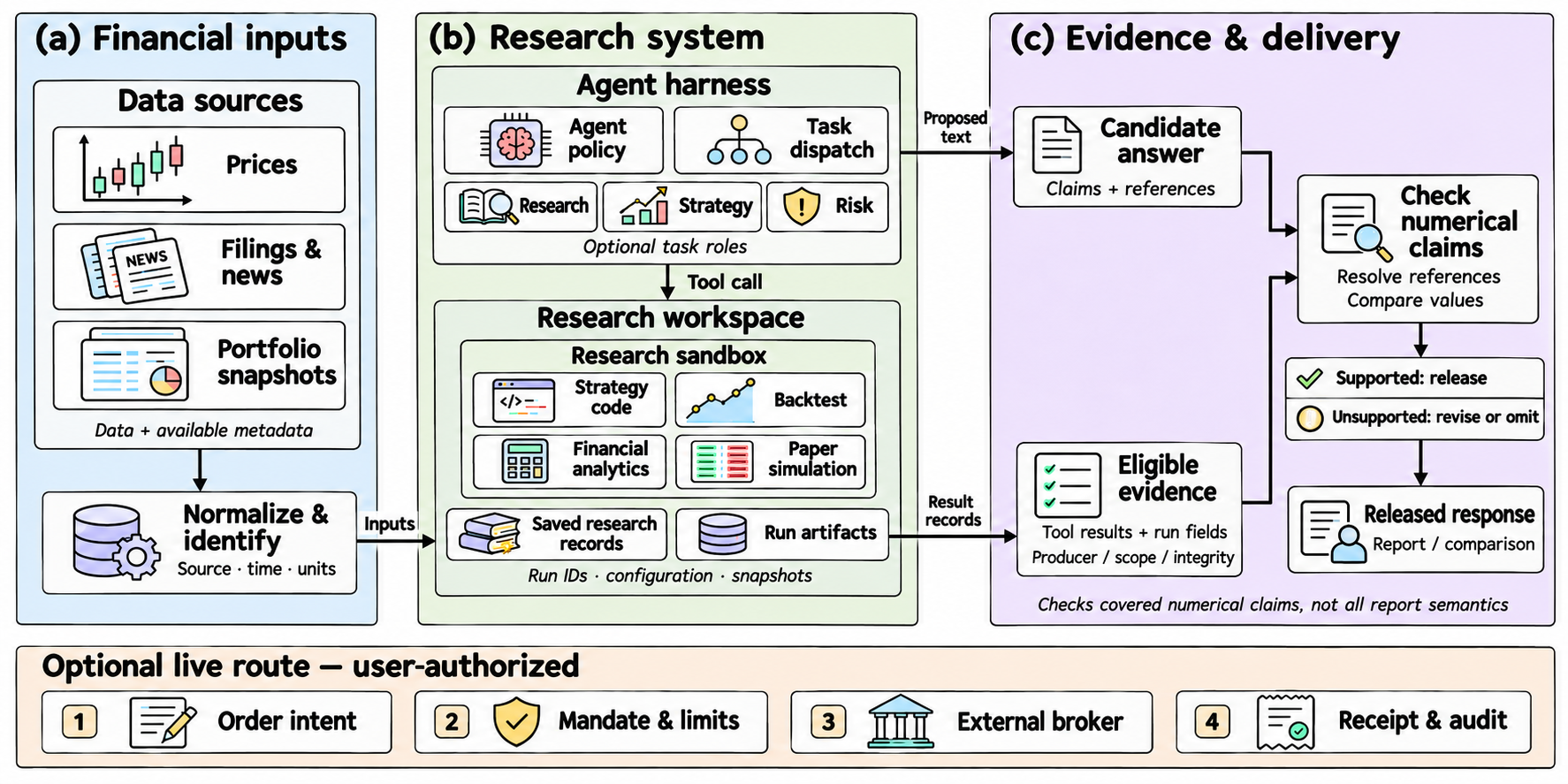}
\caption{Overview of \system{}. The workspace provides financial computation, research records and authorized execution; the harness coordinates policy decisions, tool execution and report validation.}
\label{fig:framework}
\end{figure}

\system{} consists of two cooperating layers (Figure~\ref{fig:framework}). The \textbf{workspace} provides the computation, persistent state and controlled operations of financial research, and returns every result with its financial meaning and producing run. The \textbf{harness} controls how the policy uses them within a task: which records it uses, which results support its claims and which actions it proposes. Results keep their identity across both layers, so every report and action traces back to its underlying results.

\subsection{Financial Research Workspace}\label{sec:workspace-model}\label{sec:workspace-execution}
The workspace provides callable services such as data access, backtesting, financial analytics, portfolio analysis, paper simulation and optional broker connections. Its design rests on three components: i) \textbf{financial computation contracts}, which give every result a well-defined financial meaning; ii) \textbf{persistent research artifacts}, which keep results identifiable over time; and iii) \textbf{authorized execution interfaces}, which allow external actions only under explicit authority.

\subsubsection{Financial computation contracts}
Two results become incomparable when their data or their measurement differ unnoticed. Financial data come from heterogeneous sources differing in currency, units, price adjustment and publication timing, and code implementing both a strategy and its simulator can change fills or costs along with it. The workspace therefore fixes everything in a backtest except the policy's strategy logic $\sigma$:
\[
r = \mathcal{E}\bigl(\sigma;\ \theta,\ D\bigr).
\]
\textbullet~\textbf{Aligned inputs $D$.} Every value carries the basis needed to interpret it: instrument, currency, units, price adjustment and the time it became available, such as the filing date of a fundamental. Information a source does not provide stays missing instead of being inferred.

\textbullet~\textbf{Shared measurement $\mathcal{E}$.} One engine computes fills, costs, settlement, accounting and metrics for every strategy, with market-specific settlement, price-band and fee rules. Valuation and risk analytics follow the same principle and refuse to compute without their required inputs.

\textbullet~\textbf{Recorded conditions $\theta$.} The universe, period, price convention, capital and costs are stored with each result, and frozen input data are verified by a hash manifest. Two runs that share $(\theta, D)$ therefore differ only in $\sigma$, and any other difference is visible before a gap is attributed to the strategy.

\subsubsection{Persistent research artifacts}\label{sec:workspace-artifacts}
A saved result can remain correct while no longer describing the present: a risk analysis of yesterday's portfolio still holds for yesterday's portfolio. Results therefore need identities that outlive the conversation and the market state that produced them. The workspace keeps three kinds of identified objects: \textbf{(i) runs}, each storing its configuration, metrics, validation output and artifacts, with content hashes for applicable files; \textbf{(ii) portfolio snapshots}, immutable records of holdings in which a source that fails to refresh is marked as an error and excluded from totals; and \textbf{(iii) strategies}, whose adapted versions record their parent but start unvalidated and must acquire their own run evidence. On these objects, \emph{reopening} by identifier returns the same object and refuses an unknown identifier rather than substituting current holdings, \emph{refreshing} creates a new observation with its own identifier, and \emph{adapting} creates a child that records its parent. Bounded readers return selected rows or metadata of an identified artifact and report an error instead of returning a guessed alternative. The harness passes the policy's choice of object to the corresponding operation (Section~\ref{sec:harness-execution}).

\subsubsection{Authorized execution interfaces}
An order changes an external account and cannot be undone by retrying, so acting without authority and acting twice are the costliest errors. The harness can only propose an order intent; the execution interface of the workspace decides whether it reaches the broker according to two rules:

1) \textbf{Authority is separate from evidence}: authority comes only from an account-bound mandate that the user grants for a broker connection, covering instruments, order size, exposure, leverage, daily activity and expiry. A gate checks every submission against this mandate and the current limits and halt state; research results and paper fills, however well supported, never extend it.\\
2) \textbf{An uncertain outcome is unresolved, not failed}: a broker may accept an order whose reply is lost, so a timeout is neither a failure nor a rejection. The interface keeps accepted, rejected and unresolved orders distinct and reconciles an unresolved order against broker evidence before any resubmission; on the Alpaca SDK route, a pending action with a client order identifier is persisted before the external call, so that the outcome can later be matched to the submitted intent.

\subsection{Financial Agent Harness}\label{sec:harness-model}\label{sec:policy-execution}
General-purpose agent runtimes already manage context, dispatch tools and bound retries. Financial research places two further demands on the harness: i) \textbf{state awareness}, since a step must use the intended version of a result while earlier runs and snapshots remain valid after the market moves; and ii) \textbf{evidence grounding}, since a reported number must be traceable to the run that produced it. The harness meets both by separating a result \emph{stored} in the workspace, one \emph{visible} in the policy's context and one \emph{admitted} into the evidence set $E_t$ used to check reported numbers.

\subsubsection{Context management}\label{sec:harness-context}
The harness constructs the policy context from stable instructions, task requirements, tool definitions, procedural skills, retrieved memory and current observations. Dynamic workspace state is maintained separately from the fixed instruction prefix. Skills, memory and compaction support procedural guidance, research continuity and context budget management, respectively.

\textbf{Skills.} Skills encode reusable procedural knowledge for financial research, including analytical methods, data-source guidance, investment perspectives and trading strategy workflows. The harness incorporates relevant skills during context construction and supports additional loading on demand, with loading events recorded in the execution trace. Skills guide the research procedure, while calculation inputs and associated assumptions are explicitly recorded in tool calls.

\textbf{Memory.} Memory supports research continuity across turns and sessions. Session memory retains messages, execution attempts, traces and partial responses, with checkpoints distinguishing interrupted attempts from completed ones. Research memory comprises the identified runs, portfolio snapshots and strategies defined in Section~\ref{sec:workspace-artifacts}, which are retrieved through their identifiers. Model-generated notes serve as context for subsequent reasoning, whereas numerical evidence is drawn from admitted tool outputs and engine artifacts with recorded producers.

\textbf{Compaction.} The harness manages context usage through bounded tool outputs and threshold-triggered compaction. Backtest summaries prioritize scalar metrics, validation results and artifact references, while identifying omitted content and the paths to complete outputs. As token usage reaches successive thresholds, older observations are replaced with concise summaries. These transformations affect the active context, while the underlying artifacts remain stored in the workspace and accessible through their references. Numerical claims are validated against admitted source evidence, and model-generated summaries serve as context for continued reasoning.

\subsubsection{Research execution}\label{sec:harness-execution}
\begin{figure}[t]
\centering
\includegraphics[width=\textwidth]{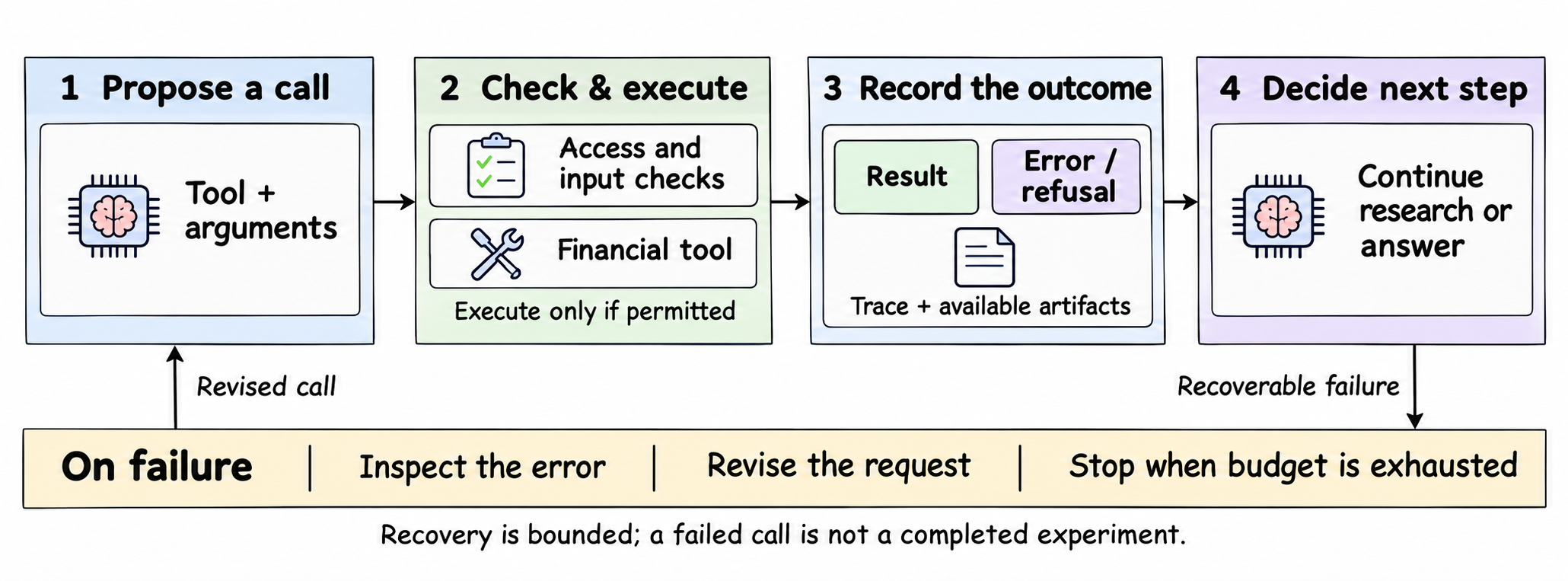}
\caption{One tool-call cycle in the harness: each policy request is executed under the relevant controls, and its result or error informs the next decision.}
\label{fig:execution}
\end{figure}

Financial research proceeds through dependent operations over persistent records and changing external state. Execution must preserve the records specified by each request and remain bounded when successive calls fail or return previously observed information. The harness connects context management with workspace operations through the following execution controls (Figure~\ref{fig:execution}).

\textbullet~\textbf{Request execution.} At each step, the policy uses the context assembled by the harness to select a workspace operation, its arguments and any relevant record identifiers. The harness dispatches the request to the corresponding service, preserving the selected run, snapshot or strategy references. The service applies the relevant access and operation checks and returns a result or an explicit error. Reopening preserves the selected record, while refreshing creates a new observation. Record selection remains the policy's responsibility, and the execution interface ensures that this selection is respected.

\textbullet~\textbf{Feedback and continuation.} Returned results and errors are recorded as observations and incorporated into the context for subsequent decisions. Tool failures provide the available diagnostic information; for file-access failures, this information distinguishes unavailable paths from access restrictions and identifies permitted alternatives. The policy uses these observations to revise its requests, while subsequent calls remain subject to the applicable tool and access constraints. The harness uses execution history to detect stalled interactions, allowing bounded recovery before terminating execution that yields no new observations with an explicit failure.

The same execution controls apply to specialist workers organized by declared task dependencies. Workers exchange saved outputs through artifact references, allowing subsequent operations to consume the identified results of earlier contributions, as stored in the workspace.

\subsubsection{Report validation}\label{sec:release-authority}
Numerical reporting requires evidence for both the reported value and its attribution to a particular observation or computation. Before release, the harness checks a candidate answer $(d,\delta)$ against the admitted evidence set $E_t$, where $d$ denotes the answer text and $\delta$ its internal numerical declarations.

\textbf{Evidence Admission.} Admission follows rules for the producing tool or engine and applicable integrity checks. Eligible tool results and engine artifacts can enter $E_t$, while files written or edited by the model are excluded as engine-produced evidence. This distinction prevents generated content from independently validating its own numerical claims, even when its contents match an engine file.

\textbf{Scoped Validation.} Each numerical declaration is checked against the call, field or run selected by its reference. A matching value from another run cannot support a claim scoped to the selected run. Derived quantities additionally require consistency between the expression and reported result, together with admitted evidence in $E_t$ for each of their data operands.

\textbf{Bounded Recovery.} Failed checks prompt reference repair, read-only evidence retrieval or draft revision by failure type and remaining budget. At the terminal boundary, eligible unsupported measurements may be omitted and the remainder revalidated; otherwise a fallback is returned.

These mechanisms implement the \textsc{Admit} and \textsc{Check} steps and the recovery branches of Algorithm~\ref{alg:task}, which summarizes how the components of this section operate within one task. Their coverage is limited to supported numerical forms and reference rules: they do not guarantee every unit label or financial interpretation, nor automatically cover report files generated by tools.

\begin{algorithm}[!htb]
\caption{One task in \system{}, from request $q$ to released answer}
\label{alg:task}
\begin{algorithmic}[1]
\Require task $q$; policy $\pi$; workspace state $s_0$; step, recovery and revision budgets
\Ensure checked or reduced answer, or a fallback; updated workspace state
\State $t \gets 0$;\ \ $E_0 \gets \emptyset$
\While{the step budget is not exhausted}
  \State $c_t \gets \mathcal{H}(q, a_{<t}, o_{<t})$ \Comment{bounded view of selected records}
  \State $a_t \sim \pi(\cdot \mid c_t)$
  \If{$a_t$ proposes an answer $(d,\delta)$}
    \State $(s_{t+1}, E_{t+1}) \gets (s_t, E_t)$
    \If{$\textsc{Check}(d,\delta,E_t)$ passes, possibly after reference repair}
      \State \Return $d$ with declarations removed
    \ElsIf{evidence recovery is within budget}
      \State admit evidence from permitted read-only calls into $E_{t+1}$
    \ElsIf{the revision budget remains}
      \State $o_t \gets$ failed checks \Comment{the policy revises its draft}
    \Else
      \State remove eligible failing measurements from $(d,\delta)$
      \State \Return $d$ if $\textsc{Check}(d,\delta,E_t)$ passes, otherwise a fallback
    \EndIf
  \Else \Comment{$a_t$ is a tool call}
    \If{$a_t$ backtests a strategy $\sigma$}
      \State $r \gets \mathcal{E}(\sigma;\theta,D)$;\ \ $o_t \gets r$;\ \ $s_{t+1} \gets s_t \cup \{(r,\theta,D)\}$
    \ElsIf{$a_t$ is an order intent outside the mandate, limits or halt state}
      \State $o_t \gets$ denial;\ \ $s_{t+1} \gets s_t$ \Comment{no submission}
    \Else
      \State $(o_t, s_{t+1}) \gets \mathcal{W}(s_t, a_t)$ \Comment{identifiers passed unchanged}
    \EndIf
    \State $E_{t+1} \gets E_t \cup \textsc{Admit}(o_t)$
    \If{recent steps added no new observation}
      \State \Return explicit failure
    \EndIf
  \EndIf
  \State $t \gets t+1$
\EndWhile
\State \Return fallback
\end{algorithmic}
\end{algorithm}

\suppressfloats[t]

%% file: sections/03_execution_evaluation.tex
\section{Experiments}\label{sec:execution}
We evaluate \system{} from request to delivered answer around four research questions. \textbf{RQ1}: Can an LLM policy complete financial research tasks through the workspace and harness, from reading its inputs to saving correct results and reports? \textbf{RQ2}: Does scoped evidence admission separate valid numerical claims from borrowed or invented values? \textbf{RQ3}: Do the controls of the workspace and harness hold at their boundaries? \textbf{RQ4}: What does evaluating computation, artifacts and delivery as separate layers reveal beyond a single numeric score?

\subsection{Experimental Setup}\label{sec:evaluation}
We evaluate with three suites: tool-use trials with a real model for RQ1 and RQ4, fixed-claim evidence controls for RQ2, and boundary regressions for RQ3. Table~\ref{tab:setup} summarizes their scale, configuration and metrics, and the following paragraphs describe each suite in detail.

\textbf{Tasks.} Two financial research templates use fixed synthetic inputs. \emph{Option pricing} values a European call and put under Black--Scholes with $S=K=100$, $T=1$, $r=0.05$, $\sigma=0.2$ and $q=0$. \emph{Historical VaR} computes the value-at-risk of ten fixed returns between $-8\%$ and $5\%$ at confidence 0.8 and horizon 1, reported as a positive loss. Each template is executed twice in a fresh workspace. We score a trial at two levels. \emph{Artifact-level success} requires reading the input, computing through the quantitative tool, and saving a numeric JSON result and a report. \emph{Delivery-level success} further requires a final answer that reports these results to the user.

\textbf{Agent configuration.} All trials run the production agent loop with DeepSeek-V4.1-Flash as the policy. Its tools read files, call the shared quantitative library and write files, and each trial is limited to twelve iterations. The model and budgets are fixed before execution, and no trial is rerun.

\begin{table}[t]
\centering\small
\caption{Evaluation suites with their research questions, scale, configuration and metrics.}
\label{tab:setup}
\begin{tabularx}{\textwidth}{@{}l c >{\raggedright\arraybackslash}p{2.1cm} Y Y@{}}
\toprule
Suite & RQ & Scale & Configuration & Metrics \\
\midrule
Tool-use trials & 1, 4 & 2 templates $\times$ 2 trials & DeepSeek-V4.1-Flash; three tools; at most 12 iterations & Numeric accuracy; verifier assertions; release outcome; time, calls and tokens \\
Evidence controls & 2 & 24 claims $\times$ 3 checks & 2 fixtures $\times$ 4 metrics; valid, wrong-run and invented claims & Agreement with the expected decision \\
Boundary regressions & 3 & 250 boundary cases, 812 subsystem checks & 9 boundary test modules; sockets disabled; isolated storage & Pass rate per suite and boundary \\
\bottomrule
\end{tabularx}
\end{table}

\textbf{Evidence controls.} Two engine-format backtest fixtures, a risk-parity run and an equal-weight run, each supply four metrics: the Sortino ratio, Calmar ratio, turnover and Sharpe ratio. For every metric and run, a valid claim cites the producing run, a wrong-run claim assigns the other run's value, and an invented claim uses a value found in neither run, giving 24 claims. Each claim is checked by the full evidence ledger and by two ablations: value-only lookup pools the values of both runs without their identity, and no artifact admission leaves engine artifacts out of the evidence set.

\textbf{Boundary regressions.} We run 250 boundary cases from nine test modules (14 November 2026) and report them alongside 812 checks across six subsystems archived from the 13 October 2026 source snapshot, all with network sockets disabled and isolated storage. Each boundary is tested through a valid control and a targeted change, as specified in Table~\ref{tab:boundaries}.

\textbf{Metrics.} Numeric accuracy compares the saved JSON fields with independent oracles, a closed-form Black--Scholes implementation and the order-statistic historical VaR, at absolute tolerance $10^{-6}$. An offline verifier asserts the integrity of each trial's saved records: the preserved request and trace, the reconciled execution status, the absence of prohibited tools, the join between tool calls and results, the requested fields and the methodology manifest, giving eight assertions per pricing trial and seven per VaR trial. We also record the release outcome of the final answer, wall-clock time, model and tool calls and token usage. Evidence controls are scored by agreement with the expected decision of each claim, and boundary regressions by their pass rate per suite and per boundary.

\subsection{End-to-End Research Execution (RQ1)}
\input{figures/eval/figA_execution}
A plausible answer alone does not show that a financial result was computed from the intended inputs or that it can be inspected afterwards. To evaluate whether \system{} supports a task from request to result, we run the complete read--compute--save--respond path with a real model on both research templates, two trials each. We score the saved results against the independent oracles, check the saved records with the offline verifier, and trace the timeline, calls and token usage of every trial. Figure~\ref{fig:rq1-execution} and Table~\ref{tab:new-evaluation} present the results, from which we make the following observations:

\begin{table}[t]
\centering\small
\caption{Per-trial outcomes and execution cost of the four tool-use trials.}
\label{tab:new-evaluation}
\begin{tabular}{@{}lccc@{\hspace{14pt}}rrrrrr@{}}
\toprule
 & \multicolumn{3}{c}{Saved outputs} & \multicolumn{6}{c}{Execution} \\
\cmidrule(lr){2-4}\cmidrule(l){5-10}
 & Abs. & Report & Verifier & Time & Model & Tool & Draft & Input & Cached \\
Trial & error & saved & assertions & (s) & calls & calls & checks & tokens & share \\
\midrule
Options 1 & 0 & $\checkmark$ & 8/8 & 56.8 & 11 & 9 & 3 & 148.7K & 84.3\% \\
Options 2 & 0 & $\checkmark$ & 8/8 & 48.8 & 10 & 9 & 4 & 134.0K & 91.1\% \\
VaR 1 & 0 & $\checkmark$ & 7/7 & 38.2 & 9 & 5 & 4 & 115.7K & 92.3\% \\
VaR 2 & 0 & $\checkmark$ & 7/7 & 34.6 & 9 & 8 & 4 & 121.4K & 92.4\% \\
\bottomrule
\end{tabular}
\end{table}

\textbullet~\textbf{Correct results through shared computation.} All four trials compute through the quantitative tool and save values identical to the oracle values, with zero absolute error: a call price of 10.45 and a put price of 5.57 in both pricing trials, and a VaR of 0.04 in both risk trials. Every trial also saves its report, and all 30 verifier assertions pass. This demonstrates that the policy obtains its results from the shared computation of the workspace, and that each result stays traceable to its request, to the tool call that produced it and to the saved record that later checks inspect.

\textbullet~\textbf{Compact execution with checked delivery.} Each trial reads its input, completes its calculations and saves its artifacts within its first 21 seconds (Figure~\ref{fig:rq1-execution}), using 5--9 tool calls and 9--11 model calls in total. The remaining time goes to three or four draft checks, which trigger evidence recovery or correction before release, and every trial finishes within 57 seconds. This indicates that the harness validates every answer before delivery while keeping a complete task within one minute.

\textbullet~\textbf{Stable context reuse.} Between 84\% and 92\% of each trial's input tokens are served from the provider's prompt cache (Table~\ref{tab:new-evaluation}), and the context grows only from about 11.5K tokens at the first call to between 14.5K and 16.8K at the last. This follows from the fixed instruction prefix of Section~\ref{sec:harness-context}: each call extends a stable context, so added checks and recoveries are mostly cached.

In summary, every trial carries its task from the input to correct saved results and a saved report, showing that all four trials reach artifact-level success through the workspace and harness, with each result produced by shared computation and retained for inspection. Delivery-level outcomes are examined in RQ4.

\subsection{Scoped Evidence Validation (RQ2)}
\input{figures/eval/figB_claims}
Numerical support has two parts: the value must appear in admitted evidence, and it must come from the run that the claim names. A check that pools values without their run confirms only the first part, and a check that excludes engine artifacts has no evidence for either. To evaluate whether scoped admission handles both, we test 24 fixed claims over two engine-format backtest runs: valid claims, wrong-run claims that borrow the other run's value, and invented claims. Each claim is checked by the full ledger and by two ablations without run identity or artifact admission (Figure~\ref{fig:rq2-claims}).

\textbf{(i)~The full ledger reaches every expected decision.} It accepts all eight valid claims and rejects all sixteen wrong-run and invented claims across four metrics and both runs. \textbf{(ii)~Run identity separates borrowed values.} Value-only lookup accepts all eight wrong-run claims, because each value exists in the pooled evidence, and only a check scoped to the declared run rejects them. \textbf{(iii)~Artifact admission enables legitimate support.} Without it, all eight valid claims are rejected even though their values come from engine outputs. Both mechanisms are therefore necessary: only their combination accepts every valid claim while rejecting every borrowed or invented value.

\subsection{Boundary Enforcement (RQ3)}
An environment for trading agents must also stop prohibited transitions, such as reading outside the permitted artifacts, replacing a selected portfolio record or submitting an order without authority, while still permitting valid work. To verify these boundaries, we specify each one by a valid control that must proceed and a targeted change that must be blocked with an observable outcome (Table~\ref{tab:boundaries}). We then run the boundary suites that exercise these pairs together with the archived checks of selected subsystems (Figure~\ref{fig:rq3-coverage}), from which we make three observations:

\begin{table}[t]
\centering\small
\caption{Workspace and harness boundaries with their test conditions and exercising suites.}
\label{tab:boundaries}
\newcommand*{\obs}[1]{\newline\textcolor{vtink2}{$\rightarrow$\,#1}\rule[-0.9ex]{0pt}{0pt}}
\newcommand*{\bnd}[2]{\textbf{#1}\newline\textbf{#2}}
\newcommand*{\sui}[1]{{\footnotesize #1}}
\newcommand*{\sub}[1]{\newline{\footnotesize\color{vtmuted}#1}}
\setlength{\tabcolsep}{5pt}\setlength{\extrarowheight}{1.5pt}
\begin{tabularx}{\textwidth}{>{\raggedright\arraybackslash}p{1.65cm}YY>{\raggedright\arraybackslash}p{3.15cm}}
\toprule
\textbf{Boundary} & \textcolor{vtgood!70!black}{$\checkmark$}\,\textbf{Valid control}\sub{condition $\rightarrow$ observation} & \textcolor{vtcrit}{$\times$}\,\textbf{Targeted change}\sub{condition $\rightarrow$ observation} & \textbf{Exercising suites}\sub{B boundary;\ S subsystem} \\
\midrule
\rowcolor{vtsurf} \bnd{Required}{inputs} & Complete valuation inputs \obs{a valuation result} & Remove a required field \obs{a named missing-input error} & \sui{S: Valuation} \\
\bnd{Artifact}{access} & Read an eligible artifact \obs{its content is returned} & Request an escaping path \obs{no outside content exposed} & \sui{B: Artifact access \newline S: Runtime and records} \\
\rowcolor{vtsurf} \bnd{Record}{selection} & Select a saved snapshot \obs{identity reaches the service} & Select an unknown identifier \obs{no holdings substituted} & \sui{B: Snapshot selection \newline S: Runtime and records} \\
\bnd{Evidence}{scope} & Cite the producing run \obs{supported claims pass} & Borrow or invent a value \obs{the claim does not pass} & \sui{B: Backtest evidence \newline S: Numerical grounding} \\
\rowcolor{vtsurf} \bnd{Trading}{authority} & Submit within the mandate \obs{the test broker receives it} & Revoke authority or halt \obs{no submission is made} & \sui{B: SDK order gate \newline S: Authority and audit} \\
\bnd{Evaluation}{coverage} & Supply the required records \obs{a verdict is produced} & Remove instrumentation \obs{a visible coverage error} & \sui{B, S: Artifact verifier} \\
\bottomrule
\end{tabularx}
\end{table}

\input{figures/eval/figC_coverage}

\textbullet~\textbf{All boundaries hold.} All 250 boundary cases and all 812 subsystem checks pass with network sockets disabled and isolated storage, and every boundary is exercised in both directions: positive controls confirm the permitted behavior, and targeted changes confirm that the prohibited transition is blocked. This shows that the controls enforce each boundary without obstructing valid work.

\textbullet~\textbf{Coverage follows consequence.} The largest suites cover evidence scope, with 95 boundary cases and 252 subsystem checks, and trading authority, with 74 boundary cases and 126 subsystem checks (Figure~\ref{fig:rq3-coverage}). These are the two boundaries where a failure would misstate a report or move funds in an external account, and their cases span producing-run scope, eligible artifact evidence, and simulated authorization and recovery paths. Testing effort thus concentrates where violations matter most.

\textbullet~\textbf{Observable oracles.} Each case inspects returned identities, artifact access or broker-call counts: a record-selection case checks the identifier that reaches the service, and an authority case counts the submissions received by the test broker. A passing case therefore shows from the state of the system that the prohibited transition did not take place. Overall, the controls of the workspace and harness hold at all six boundaries of Table~\ref{tab:boundaries}, for both valid controls and targeted changes.

\subsection{Layered Diagnosis of Delivery (RQ4)}
\input{figures/eval/figD_layers}
A numeric score shows whether a saved value is correct, but the user receives a report and a reply, whose meaning and completeness can differ from the saved numbers. To examine what separate layers reveal, we re-evaluate the four tool-use trials of RQ1 from the numeric result through the saved report and the artifact assertions to the final release, and inspect the meaning of each report and reply separately from its numbers. Figure~\ref{fig:evaluation} presents the results, from which we make two observations:

\textbullet~\textbf{Completion differs across layers.} Numeric results, saved reports and artifact assertions pass in all four trials (Figure~\ref{fig:evaluation}a), so a score based on numeric artifacts alone would rate every trial as complete. The release layer separates the outcomes: one trial releases its full answer, two release answers with unsupported figures removed, and one declines to confirm results it had correctly saved (Figure~\ref{fig:evaluation}d). This demonstrates that the layer closest to the user carries information that the lower layers do not.

\textbullet~\textbf{Each layer exposes a distinct condition.} Both pricing reports describe the tool's per-day theta as an annual figure (Figure~\ref{fig:evaluation}c): the number is correct while its unit is not, which only an inspection of report semantics reveals. A calculation blocked by a missing required input (Figure~\ref{fig:evaluation}b) is the intended behavior, whereas a refusal after a correct calculation is not, and separate layers keep the two apart. In summary, layered evaluation locates where completion changes between computation and delivery, motivating the extensions to evaluation and checking in Section~\ref{sec:insights}.

%% file: figures/eval/figA_execution.tex
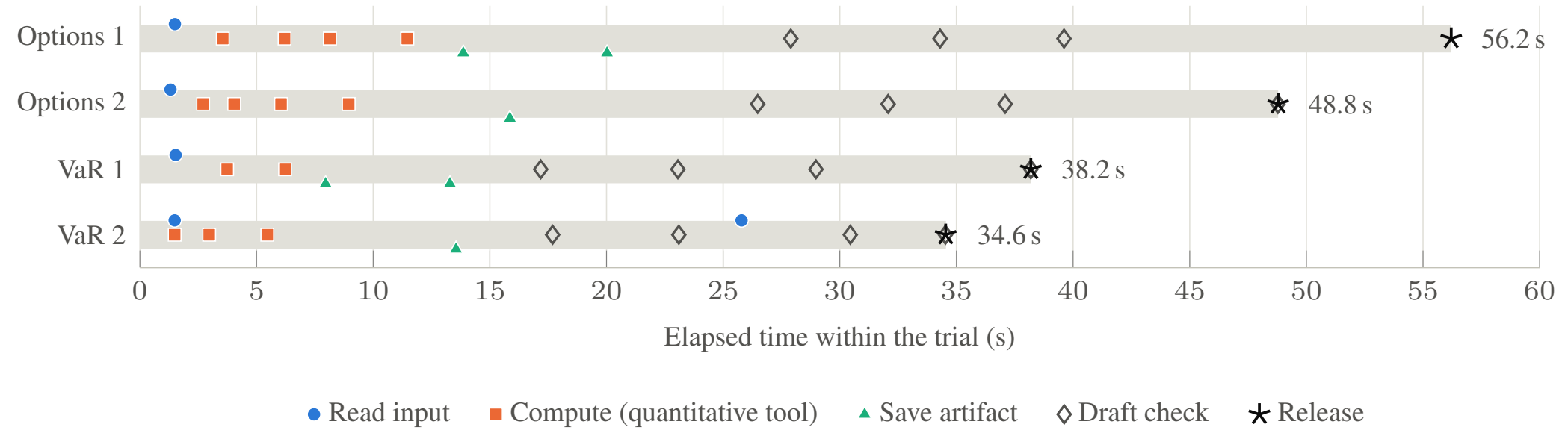
\begin{figure}[t]
\centering
\begin{tikzpicture}
\begin{axis}[vtaxis,
  width=\linewidth, height=3.9cm,
  xmin=0, xmax=60, ymin=0.5, ymax=4.5,
  ytick={1,2,3,4}, yticklabels={VaR 2, VaR 1, Options 2, Options 1},
  xlabel={Elapsed time within the trial (s)},
  xmajorgrids, axis x line*=bottom, axis y line*=left, y axis line style={draw=none},
  ytick style={draw=none},
  legend columns=5, legend style={at={(0.5,-0.45)}, anchor=north, /tikz/every even column/.append style={column sep=8pt}},
  clip=false]
  \addplot[vtgrid, line width=7pt, forget plot] coordinates {(0,4) (56.21,4)};
  \addplot[vtgrid, line width=7pt, forget plot] coordinates {(0,3) (48.79,3)};
  \addplot[vtgrid, line width=7pt, forget plot] coordinates {(0,2) (38.19,2)};
  \addplot[vtgrid, line width=7pt, forget plot] coordinates {(0,1) (34.55,1)};
  \addplot[only marks, mark=*, mark size=1.7pt, vtblue, mark options={draw=white, line width=0.3pt}]
    table[x=x, y expr=\thisrow{y}+0.22] {figures/eval/data/timeline_read.dat};
  \addlegendentry{Read input}
  \addplot[only marks, mark=square*, mark size=1.6pt, vtorange, mark options={draw=white, line width=0.3pt}]
    table[x=x, y=y] {figures/eval/data/timeline_compute.dat};
  \addlegendentry{Compute (quantitative tool)}
  \addplot[only marks, mark=triangle*, mark size=2.0pt, vtaqua, mark options={draw=white, line width=0.3pt}]
    table[x=x, y expr=\thisrow{y}-0.22] {figures/eval/data/timeline_write.dat};
  \addlegendentry{Save artifact}
  \addplot[only marks, mark=diamond, mark size=2.2pt, vtink2, line width=0.6pt]
    table[x=x, y=y] {figures/eval/data/timeline_check.dat};
  \addlegendentry{Draft check}
  \addplot[only marks, mark=star, mark size=2.8pt, vtink, line width=0.7pt]
    table[x=x, y=y] {figures/eval/data/timeline_release.dat};
  \addlegendentry{Release}
  \node[font=\scriptsize, text=vtink2, anchor=west] at (axis cs:56.9,4) {56.2\,s};
  \node[font=\scriptsize, text=vtink2, anchor=west] at (axis cs:49.5,3) {48.8\,s};
  \node[font=\scriptsize, text=vtink2, anchor=west] at (axis cs:38.9,2) {38.2\,s};
  \node[font=\scriptsize, text=vtink2, anchor=west] at (axis cs:35.3,1) {34.6\,s};
\end{axis}
\end{tikzpicture}
\caption{Execution timelines of the four tool-use trials.}
\label{fig:rq1-execution}
\end{figure}

%% file: figures/eval/figB_claims.tex
\input{figures/eval/data/claims_cells.tex}
\begin{figure}[t]
\centering
\begin{tikzpicture}[x=1cm, y=1cm]
  \def\cw{0.82}\def\ch{0.40}\def\gap{0.22}
  \foreach \g/\lab/\exp in {0/Valid claims/accept, 1/Wrong-run claims/reject, 2/Invented claims/reject}{
    \pgfmathsetmacro{\gx}{\g*(3*\cw+\gap)}
    \node[font=\scriptsize\bfseries, text=vtink, anchor=south] at (\gx+1.5*\cw, 8*\ch+0.74) {\lab};
    \node[font=\tiny, text=vtmuted, anchor=south] at (\gx+1.5*\cw, 8*\ch+0.53) {expected: \exp};
    \draw[vtaxis, line width=0.4pt] (\gx+0.04, 8*\ch+0.52) -- (\gx+3*\cw-0.04, 8*\ch+0.52);
    \foreach \a/\alab in {0/{Full\\ledger}, 1/{Value\\only}, 2/{No\\artifact}}{
      \node[font=\tiny, text=vtink2, anchor=south, align=center, inner sep=1pt] at (\gx+\a*\cw+0.5*\cw, 8*\ch+0.03) {\alab};
    }
  }
  \foreach \y/\lab in {8/Sortino, 7/Calmar, 6/Turnover, 5/Sharpe, 4/Sortino, 3/Calmar, 2/Turnover, 1/Sharpe}{
    \node[font=\scriptsize, text=vtink2, anchor=east] at (-0.08, \y*\ch-0.5*\ch) {\lab};
  }
  \node[font=\scriptsize\bfseries, text=vtink, rotate=90] at (-1.38, 6*\ch) {Risk parity};
  \node[font=\scriptsize\bfseries, text=vtink, rotate=90] at (-1.38, 2*\ch) {Equal weight};
  \foreach \x/\y/\acc/\ok in \claimcells{
    \pgfmathsetmacro{\px}{(\x-1)*\cw + int((\x-1)/3)*\gap}
    \pgfmathsetmacro{\py}{(\y-1)*\ch}
    \ifnum\ok=1 \def\cellfill{vtgood!16}\def\glyphcol{vtgood!70!black}\else \def\cellfill{vtcrit!22}\def\glyphcol{vtcrit}\fi
    \fill[\cellfill, rounded corners=1pt] (\px+0.025, \py+0.025) rectangle (\px+\cw-0.025, \py+\ch-0.025);
    \ifnum\acc=1
      \node[font=\scriptsize, text=\glyphcol] at (\px+0.5*\cw, \py+0.5*\ch) {$\checkmark$};
    \else
      \node[font=\scriptsize, text=\glyphcol] at (\px+0.5*\cw, \py+0.5*\ch) {$\times$};
    \fi
  }
  \draw[vtaxis, line width=0.4pt] (-1.05, 4*\ch) -- (9*\cw+2*\gap, 4*\ch);
  \begin{scope}[shift={(0, -0.36)}]
    \fill[vtgood!16, rounded corners=1pt] (0,0) rectangle (0.30,0.20); \node[font=\tiny, text=vtink2, anchor=west] at (0.34,0.10) {matches expectation};
    \fill[vtcrit!22, rounded corners=1pt] (2.40,0) rectangle (2.70,0.20); \node[font=\tiny, text=vtink2, anchor=west] at (2.74,0.10) {contradicts expectation};
    \node[font=\tiny, text=vtink2, anchor=west] at (5.25,0.10) {$\checkmark$ accepted \quad $\times$ rejected};
  \end{scope}
  \begin{axis}[vtaxis, name=acc, at={(9.40cm,0.29cm)}, anchor=south west,
    scale only axis, width=2.7cm, height=8*0.40cm,
    xbar stacked, /pgf/bar width=13pt,
    xmin=0, xmax=24, xtick={0,8,16,24},
    ytick={1,2,3}, yticklabels={No artifact, Value-only, Full ledger},
    ymin=0.45, ymax=3.55,
    axis x line*=bottom, axis y line*=left, y axis line style={draw=none}, ytick style={draw=none},
    xmajorgrids, xlabel={Correct decisions (of 24)}, xlabel style={yshift=2pt},
    clip=false]
    \addplot[fill=vtgood!42, draw=white, line width=0.6pt] coordinates {(16,1) (16,2) (24,3)};
    \addplot[fill=vtcrit!42, draw=white, line width=0.6pt] coordinates {(8,1) (8,2) (0,3)};
    \coordinate (s3) at (axis cs:0.4,3); \coordinate (s2) at (axis cs:0.4,2); \coordinate (s1) at (axis cs:0.4,1);
    \coordinate (e2) at (axis cs:20,2); \coordinate (e1) at (axis cs:20,1);
    \coordinate (mid) at (axis cs:12,0.45);
  \end{axis}
  \coordinate (ysub) at (0,-0.55);
  \node[subcap, anchor=north] at (3.70, -0.55) {(a) Decision on each claim};
  \node[subcap, anchor=north] at (mid |- ysub) {(b) Accuracy by check};
  \node[font=\scriptsize\bfseries, text=vtink, anchor=west] at (s3) {24/24};
  \node[font=\scriptsize\bfseries, text=vtink, anchor=west] at (s2) {16/24};
  \node[font=\scriptsize\bfseries, text=vtink, anchor=west] at (s1) {16/24};
  \node[font=\tiny, text=vtink, align=center] at (e2) {wrong\\run};
  \node[font=\tiny, text=vtink, align=center] at (e1) {valid};
\end{tikzpicture}
\caption{Decisions of the full ledger and its two ablations on the 24 fixed claims.}
\label{fig:rq2-claims}
\end{figure}
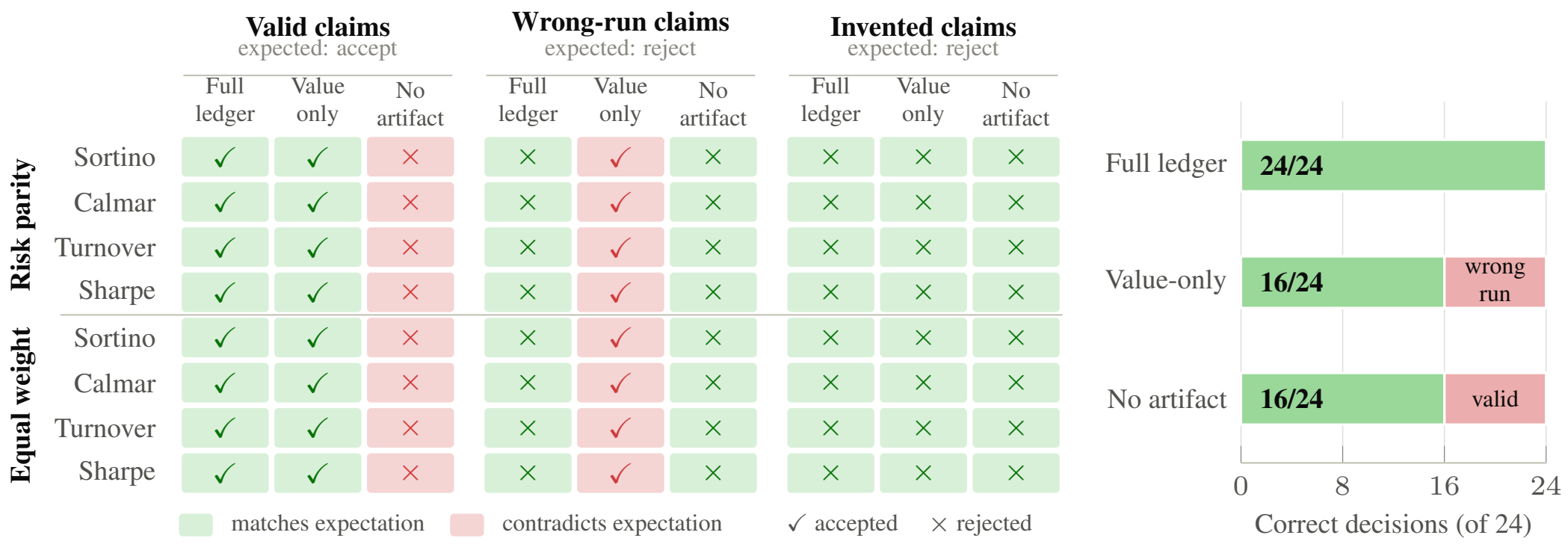

%% file: figures/eval/figC_coverage.tex
\begin{figure}[t]
\centering
\begin{tikzpicture}
\begin{groupplot}[vtaxis,
  group style={group size=2 by 1, horizontal sep=3.1cm},
  width=0.36\linewidth, height=3.7cm,
  xbar, /pgf/bar width=7pt,
  xmin=0, enlarge y limits=0.12,
  axis x line*=bottom, axis y line*=left, y axis line style={draw=none}, ytick style={draw=none},
  xmajorgrids, clip=false,
  nodes near coords, nodes near coords style={font=\scriptsize, text=vtink, anchor=west},
  every node near coord/.append style={xshift=1pt}]
\nextgroupplot[
  xmax=110, xtick={0,25,50,75,100}, xlabel={Test cases},
  ytick={1,2,3,4,5,6},
  yticklabels={Snapshot selection, Artifact access, Quantitative tool, Artifact verifier, SDK order gate, Backtest evidence}]
  \addplot[fill=vtblue, draw=none] coordinates {(1,1) (7,2) (27,3) (46,4) (74,5) (95,6)};
\nextgroupplot[
  xmax=290, xtick={0,100,200}, xlabel={Test cases},
  ytick={1,2,3,4,5,6},
  yticklabels={Artifact verifier, Runtime and records, Authority and audit, Data and factors, Valuation, Numerical grounding}]
  \addplot[fill=vtblueD, draw=none] coordinates {(21,1) (83,2) (126,3) (137,4) (193,5) (252,6)};
\end{groupplot}
\node[subcap, anchor=north] at ($(group c1r1.outer south)+(0,-2pt)$) {(a) Boundary suites: 250/250 passed};
\node[subcap, anchor=north] at ($(group c2r1.outer south)+(0,-2pt)$) {(b) Subsystem checks: 812/812 passed};
\end{tikzpicture}
\caption{Regression cases per boundary suite and per subsystem.}
\label{fig:rq3-coverage}
\end{figure}
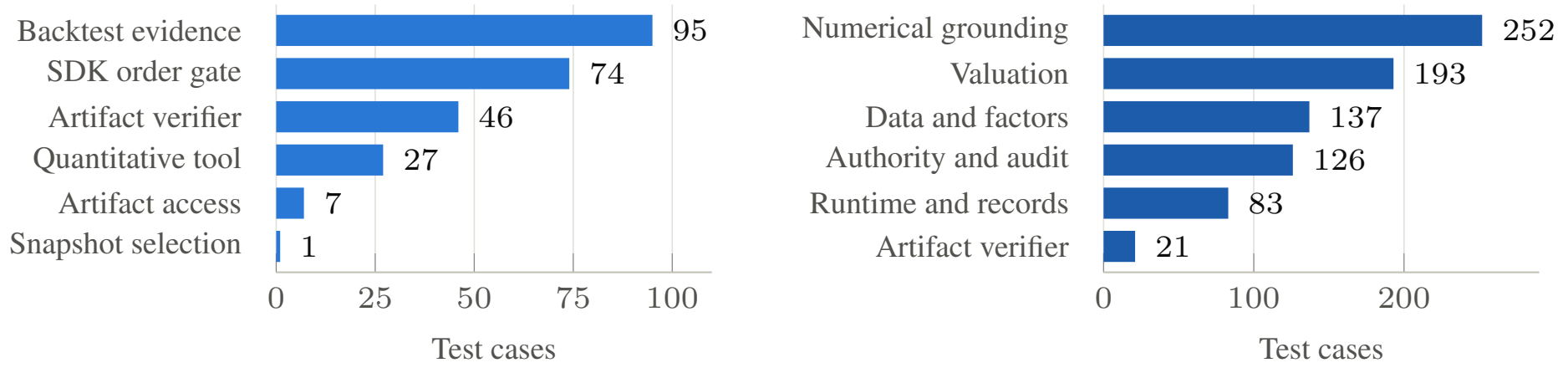

%% file: figures/eval/figD_layers.tex
\begin{figure}[t]
\centering
\begin{tikzpicture}[x=1cm, y=1cm,
  cell/.style={minimum width=0.86cm, minimum height=0.48cm, inner sep=0pt, font=\tiny, text=vtink, rounded corners=1pt},
  good/.style={cell, fill=vtgood!22},
  warn/.style={cell, fill=vtwarn!38},
  crit/.style={cell, fill=vtcrit!25}]
  \foreach \c/\lab in {0/Options 1, 1/Options 2, 2/VaR 1, 3/VaR 2}{
    \node[font=\tiny, text=vtink2] at (\c*0.92, 3.15) {\lab};
  }
  \foreach \r/\lab in {3/Numeric result, 2/Saved report, 1/Artifact assertions, 0/Final release}{
    \node[font=\scriptsize, text=vtink2, anchor=east] at (-0.52, \r*0.68+0.55) {\lab};
  }
  \foreach \c in {0,1,2,3}{
    \node[good] at (\c*0.92, 3*0.68+0.55) {$\checkmark$ exact};
    \node[good] at (\c*0.92, 2*0.68+0.55) {$\checkmark$ saved};
  }
  \node[good] at (0*0.92, 1*0.68+0.55) {$\checkmark$ 8/8};
  \node[good] at (1*0.92, 1*0.68+0.55) {$\checkmark$ 8/8};
  \node[good] at (2*0.92, 1*0.68+0.55) {$\checkmark$ 7/7};
  \node[good] at (3*0.92, 1*0.68+0.55) {$\checkmark$ 7/7};
  \node[good] at (0*0.92, 0.55) {$\checkmark$ full};
  \node[crit] at (1*0.92, 0.55) {$\times$ refused};
  \node[warn] at (2*0.92, 0.55) {$\circ$ reduced};
  \node[warn] at (3*0.92, 0.55) {$\circ$ reduced};
  \draw[vtaxis, line width=0.4pt] (-2.45, 0.89) -- (3.32, 0.89);
  \node[font=\tiny, text=vtmuted, anchor=north west] at (-2.45, 0.24)
    {reduced: released with unsupported figures removed};
  \node[subcap, anchor=north] at (0.43, -0.18) {(a) Outcome by evaluation layer};
  \def\hl#1{{\setlength{\fboxsep}{0.9pt}\colorbox{vtwarn!38}{#1}}}
  \begin{scope}[shift={(4.05,0)},
    box/.style={draw=vtaxis, fill=vtsurf, rounded corners=2pt, font=\tiny, text=vtink, align=left, text width=3.05cm, minimum height=0.66cm, inner sep=3pt},
    arr/.style={-{Stealth[length=3pt]}, vtmuted, line width=0.6pt}]
    \node[box, anchor=west] (b1) at (0, 2.88) {Valuation request\\required cash flow: \hl{missing}};
    \node[box, anchor=west] (b2) at (3.85, 2.88) {Service response\\missing-input error, no result};
    \draw[arr] (b1) -- (b2);
    \node[subcap, anchor=north] at (3.55, 2.46) {(b) Blocked input \textcolor{vtmuted}{(illustrative)}};
    \node[box, anchor=west] (c1) at (0, 1.56) {Tool output, per calendar day\\$\theta=-0.0176$};
    \node[box, anchor=west] (c2) at (3.85, 1.56) {Saved report, \hl{per year}\\$\theta=-0.0176$};
    \draw[arr] (c1) -- (c2);
    \node[subcap, anchor=north] at (3.55, 1.14) {(c) Report semantics \textcolor{vtmuted}{(Options 1)}};
    \node[box, anchor=west] (d1) at (0, 0.24) {Saved result\\call 10.45, put 5.57};
    \node[box, anchor=west] (d2) at (3.85, 0.24) {Final reply (paraphrased)\\\hl{prices could not be verified}};
    \draw[arr] (d1) -- (d2);
    \node[subcap, anchor=north] at (3.55, -0.18) {(d) Delivery \textcolor{vtmuted}{(Options 2)}};
  \end{scope}
\end{tikzpicture}
\caption{Layered evaluation of the four tool-use trials and three representative cases.}
\label{fig:evaluation}
\end{figure}
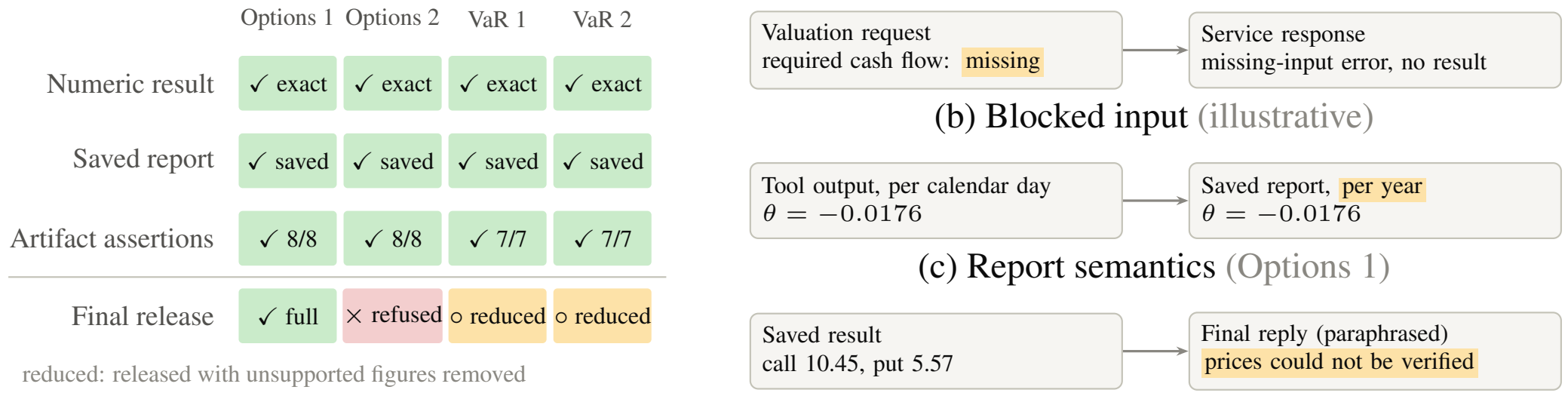

%% file: sections/04_insights.tex
\section{Discussion}\label{sec:insights}
Our results point to two insights for building trading agents and two challenges for the next steps.

\subsection{Environment and strategies as the key lever}
Current LLMs already plan and decide well within a task. In every trial, the policy selected the appropriate tool and reached exact results within a few calls, and the differences that remained concerned interpretation and delivery, which the environment can make explicit. For trading agents, the decisive factor is what surrounds the policy: reliable data and computation with explicit conventions, records that preserve earlier work, and a comprehensive repertoire of strategies and analysis methods from which the policy can choose. \system{} keeps the policy replaceable and builds these components, so that advances in model capability carry over to research outcomes.

\subsection{Assuring results through the environment}
An agent's result is useful only when it can be relied on: computed from the intended inputs, reported with the correct meaning and delivered as requested. The cases in Section~\ref{sec:execution} show that this assurance is a property of the environment and involves three requirements. \textbf{(i)~Attribution with meaning.} Run identity separates values by their source (Figure~\ref{fig:rq2-claims}), while the theta case shows that a correctly attributed value can still carry the wrong unit or time basis (Figure~\ref{fig:evaluation}c); checks should therefore attach machine-readable measure and unit conventions to each result and verify them against the report claim~\citep{gao2023alce,gao2023rarr}. \textbf{(ii)~Delivery as part of the result.} The offline verifier confirms stored results, while the release layer shows that delivery can still differ (Figure~\ref{fig:evaluation}); task specifications should state what must reach the user, record refusals and omissions explicitly, and credit justified refusals such as a blocked missing input. \textbf{(iii)~Comparable evaluation conditions.} Policy comparisons hold when data access, tools, conventions and budgets are shared and execution coverage is reported; since completion can depend on the policy, all assigned tasks should remain in completion statistics, with quality compared on a common completed subset whose size and denominator are reported.

\subsection{Open challenges}
\textbullet~\textbf{From open-ended goals to strategies.} Users often state an outcome without a method, as in ``earn 2\% from the stock market today.'' Serving such a request requires the agent to translate the goal into candidate strategies, estimate their chance of success under current market conditions, and select the strategy with the highest expected win rate while stating the risk that the goal is not met. Because markets move and repeated search inflates apparent performance, this selection must account for regime changes as well as selection bias and backtest overfitting~\citep{white2000realitycheck,bailey2016backtestoverfitting,harvey2016crosssection}.

\textbullet~\textbf{Self-evolution across tasks.} An agent that runs many tasks accumulates evidence about which strategies worked under which conditions. Turning this experience into better later decisions requires deciding what to retain, recognizing when past results no longer apply as the market shifts, and verifying that each update improves outcomes. The persistent runs, snapshots and strategies of the workspace (Section~\ref{sec:workspace-artifacts}) provide the substrate for such self-evolution, which remains to be studied with matched, task-level experiments over changing market conditions.

%% file: sections/05_related_work.tex
\section{Related Work}\label{sec:related}

\subsection{LLM Agents}
LLM agents extend language models to interactive problem solving by reasoning over observations and acting through tools~\citep{wang2024agentsurvey}. Early approaches interleave reasoning with actions, learn API calls, or use executable code as a unified action space~\citep{yao2023react,schick2023toolformer,qin2024toolllm,wang2024codeact}, and later work adds self-reflection, tree search and memory for long-horizon tasks~\citep{shinn2023reflexion,zhou2024lats,park2023generative,packer2023memgpt,wang2023voyager,wang2025awm}. Agents for software engineering support repository understanding and code development through tool interfaces, context management, and execution feedback~\citep{yang2024sweagent,wang2025openhands,li2026fastcode,li2025deepcode}. Multi-agent frameworks organize collaborative work through role assignment, communication, and coordination of intermediate outputs~\citep{wu2023autogen,hong2024metagpt,li2026autoswarm}. However, financial research further requires data, computations and records to preserve their conventions, timing and provenance across steps, which the environment around the agent must supply.

\subsection{Agentic Financial Systems}
LLMs entered finance through domain-adapted models trained on financial corpora~\citep{wu2023bloomberggpt,yang2023fingpt}. Single-agent traders add layered memory, multimodal tools with reflection, or policy-gradient fine-tuning of the LLM~\citep{yu2023finmem,zhang2024finagent,xiong2025flagtrader}, while multi-agent systems mirror investment firms through debating roles, manager--analyst hierarchies and simulated trading~\citep{xiao2024tradingagents,yu2024fincon,li2025quantagents,yang2024finrobot}. Research agents automate factor and model development~\citep{li2025rdagentquant}, unified toolchains standardize market data and simulated brokerage~\citep{luo2026qfinzero}, and benchmarks evaluate agents from financial tasks to live trading and research~\citep{xie2024finben,li2025investorbench,li2025aitrader,bigeard2025financeagent}. While these systems advance the agent's decision logic, building environments that keep its results correct, attributable and reusable across tasks remains an important direction.

%% file: sections/05_conclusion.tex
\section{Conclusion}\label{sec:conclusion}
We presented \system{}, an agent-native environment that provides LLMs with a financial research workspace and an execution harness. The workspace provides shared financial computation, persistent research records, and authorized execution, preserving the provenance and execution conditions needed to interpret and reuse research results. Building on these resources, the harness coordinates context management, research execution with bounded recovery, and report validation against admitted evidence. Together, the two components enable LLMs to automate financial research end to end, from data access and quantitative computation to reported findings, while keeping each result reliable and traceable to the computation and evidence that support it. Task evaluations with LLMs, controlled evidence comparisons, and regression tests provide empirical support for these mechanisms while clarifying the framework's operational scope and limitations.

%% file: sections/appendix_evidence.tex
\refstepcounter{section}\label{app:evidence}
\section*{Appendix}

\subsection{Evaluation Package}\label{app:package}
The evaluation package retains the frozen task inputs and prompts, the execution and scoring scripts, per-trial traces and token usage, final responses, generated artifacts, source fingerprints, the claim matrix and the JUnit records of the boundary suites. The subsystem checks retain their own commands, module lists, source fingerprints, raw logs and JUnit files. All evaluation figures are drawn as vector graphics directly from these records, and reproducing the recorded results requires the matching code version and dependencies together with the retained commands and fingerprints.

\subsection{Tool-Use Trials}\label{app:trials}
The pricing oracle evaluates the Black--Scholes formula through the error function, giving call and put prices of 10.45 and 5.57. The VaR oracle sorts the returns $[-.08,-.04,-.02,-.01,0,.01,.02,.03,.04,.05]$ and takes the second observation without interpolation, giving a positive loss of 0.04. The registry offers no broker or market-data tool, and trials invoke the production agent loop directly. Source fingerprints were captured during the first trial and matched again after all four, so every trial ran on the same code version.

\input{figures/eval/figE_trials}

Figure~\ref{fig:app-trials}a lists the verifier assertions of each trial: eight for each pricing trial and seven for each VaR trial, which has no second requested field. Figure~\ref{fig:app-trials}b counts the harness events before release, together with each trial's release outcome; the trials produce between 4.5K and 8.2K output tokens. Report semantics and the final reply are assessed separately from these assertions: unit labels are checked against the tool's documented return convention, and all outputs are preserved unrepaired.

\subsection{Evidence-Control Claims}\label{app:claims}
Figure~\ref{fig:app-claims} plots the claimed values of the 24 claims. Each wrong-run claim takes the value of the same metric in the other run, so it coincides with a valid value of that run (dotted links), while each invented claim lies apart from the values of both runs. The check without artifact admission uses an empty ledger, and value-only lookup pools the numerical values of both runs at the same displayed precision. The full claim matrix and the decision reasons are retained for all 24 comparisons.

\input{figures/eval/figF_claimvalues}

\subsection{Regression Inventory}\label{app:regressions}
Table~\ref{tab:regressions} lists the boundary suites and subsystem checks with the scope that each exercises. Counts are parsed from the retained JUnit files (boundary suites: 14 November 2026; subsystem checks: 13 October 2026) and include parameterized cases, bars share one scale across both inventories, and no suite records a failure, an error or a skip.

\begin{table}[h]
\centering\small
\caption{Boundary suites and subsystem checks with their exercised scope.}
\label{tab:regressions}
\newcommand*{\spark}[2]{\textcolor{#2}{\rule{#1}{1.3ex}}}
\newcommand*{\grp}[2]{\rowcolor{vtsurf}\multicolumn{4}{@{\hspace{5pt}}l}{\textbf{#1}\quad{\footnotesize\color{vtink2}#2}}}
\setlength{\tabcolsep}{5pt}\setlength{\extrarowheight}{1pt}
\begin{tabularx}{\textwidth}{>{\raggedright\arraybackslash}p{2.9cm}r>{\raggedright\arraybackslash}p{1.45cm}Y}
\toprule
Suite & Cases & & Exercised scope \\
\midrule
\grp{Boundary suites}{250/250 passed} \\
Backtest evidence & 95 & \spark{0.528cm}{vtblue} & Producing-run scope and eligible artifact evidence \\
SDK order gate & 74 & \spark{0.411cm}{vtblue} & Simulated authorization and recovery paths \\
Artifact verifier & 46 & \spark{0.256cm}{vtblue} & Cases, persisted records and verdicts \\
Quantitative tool & 27 & \spark{0.150cm}{vtblue} & Allowed calls and refused functions \\
Artifact access & 7 & \spark{0.039cm}{vtblue} & Read refusal and directory errors \\
Snapshot selection & 1 & \spark{0.02cm}{vtblue} & Requested identifier reaches the service \\
\grp{Subsystem checks}{812/812 passed} \\
Numerical grounding & 252 & \spark{1.400cm}{vtblueD} & Operand and run evidence, release behavior, numeric precision and check registration \\
Valuation & 193 & \spark{1.072cm}{vtblueD} & Missing-input contracts, valuation models, linked accounting and export artifacts \\
Data and factors & 137 & \spark{0.761cm}{vtblueD} & Filing availability, lag, routing, annualization and selected factor calculations \\
Authority and audit & 126 & \spark{0.700cm}{vtblueD} & SDK gate paths, simulated recovery, paper-capped refusal and governance primitives \\
Runtime and records & 83 & \spark{0.461cm}{vtblueD} & Context budget, snapshot identifier propagation, artifact access and manifest wiring \\
Artifact verifier & 21 & \spark{0.117cm}{vtblueD} & Case validation, persisted-run parsing and verdict production \\
\bottomrule
\end{tabularx}
\end{table}

\subsection{Deployment Scope and Disclosure}\label{app:disclosure}
\paragraph{Deployment scope.}
The implementation enforces application-level contracts on supported paths. Hashes establish consistency with recorded bytes, but do not authenticate provider truth or creation time. Local checkpoints and locks do not coordinate every external client. Deployments permitting unrestricted shell access under the same operating-system identity require host-level filesystem, process and credential isolation in addition to these application-level contracts~\citep{saltzer1975protection,greshake2023indirect}.

\paragraph{Disclosure.}
AI assistance was used for manuscript preparation and raster figure generation. Figures~\ref{fig:framework} and~\ref{fig:execution} illustrate architecture and execution; the evaluation figures derive from retained records. Figure~\ref{fig:evaluation}b is schematic, and Figure~\ref{fig:evaluation}c--d abbreviate retained outputs. The two-run fixtures of the evidence controls (Figure~\ref{fig:rq2-claims}) are synthetic. Figure-generation prompts and asset hashes accompany the manuscript; borrowed visual inspiration implies no reuse of another system's design or results.

%% file: figures/eval/figE_trials.tex
\begin{figure}[h]
\centering
\begin{tikzpicture}[x=1cm, y=1cm,
  ok/.style={fill=vtgood!20, rounded corners=1pt},
  na/.style={fill=vtsurf, rounded corners=1pt}]
  \def\cw{0.88}\def\ch{0.36}
  \foreach \c/\lab in {0/Options 1, 1/Options 2, 2/VaR 1, 3/VaR 2}{
    \node[font=\tiny, text=vtink2, anchor=south] at (\c*\cw+0.5*\cw, 8*\ch+0.02) {\lab};
  }
  \foreach \r/\lab in {8/Request preserved, 7/Trace preserved, 6/Execution status reconciled, 5/No prohibited tool,
                       4/Tool calls joined to results, 3/Requested field: call or VaR, 2/Requested field: put, 1/Methodology manifest intact}{
    \node[font=\scriptsize, text=vtink2, anchor=east] at (-0.08, \r*\ch-0.5*\ch) {\lab};
    \foreach \c in {0,1,2,3}{
      \pgfmathtruncatemacro{\isna}{(\r==2) && (\c>1)}
      \ifnum\isna=1
        \fill[na] (\c*\cw+0.03, \r*\ch-\ch+0.03) rectangle (\c*\cw+\cw-0.03, \r*\ch-0.03);
        \node[font=\tiny, text=vtmuted] at (\c*\cw+0.5*\cw, \r*\ch-0.5*\ch) {n/a};
      \else
        \fill[ok] (\c*\cw+0.03, \r*\ch-\ch+0.03) rectangle (\c*\cw+\cw-0.03, \r*\ch-0.03);
        \node[font=\scriptsize, text=vtgood!70!black] at (\c*\cw+0.5*\cw, \r*\ch-0.5*\ch) {$\checkmark$};
      \fi
    }
  }
  \draw[vtaxis, line width=0.4pt] (-3.35, -0.04) -- (4*\cw, -0.04);
  \node[font=\scriptsize\bfseries, text=vtink, anchor=east] at (-0.08, -0.26) {Passed};
  \foreach \c/\v in {0/{8\,/\,8}, 1/{8\,/\,8}, 2/{7\,/\,7}, 3/{7\,/\,7}}{
    \node[font=\scriptsize\bfseries, text=vtink] at (\c*\cw+0.5*\cw, -0.26) {\v};
  }
  \node[subcap, anchor=north] at (0.30, -0.80) {(a) Verifier assertions};
  \begin{axis}[vtaxis, name=ev, at={(5.00cm,0cm)}, anchor=south west,
    scale only axis, width=4.95cm, height=8*0.36cm,
    ybar, /pgf/bar width=6.5pt, bar shift auto,
    ymin=0, ymax=5, ytick={0,1,2,3,4}, ylabel={Events per trial},
    xmin=0.45, xmax=4.55, xtick={1,2,3,4},
    xticklabels={{Options 1\\\textcolor{vtgood!70!black}{$\checkmark$}\,full},
                 {Options 2\\\textcolor{vtcrit}{$\times$}\,refused},
                 {VaR 1\\\textcolor{vtwarn!85!black}{$\circ$}\,reduced},
                 {VaR 2\\\textcolor{vtwarn!85!black}{$\circ$}\,reduced}},
    xticklabel style={align=center, font=\tiny, text=vtink2},
    ymajorgrids, axis x line*=bottom, axis y line*=left, xtick style={draw=none}, ytick style={draw=none},
    nodes near coords, nodes near coords style={font=\tiny, text=vtink2},
    legend columns=3, legend style={at={(0.5,1.02)}, anchor=south, /tikz/every even column/.append style={column sep=4pt}},
    legend image code/.code={\fill[#1] (0cm,-0.07cm) rectangle (0.22cm,0.09cm);},
    clip=false]
    \addplot[fill=vtblue, draw=none] coordinates {(1,3) (2,4) (3,4) (4,4)};
    \addlegendentry{Draft checks}
    \addplot[fill=vtorange, draw=none] coordinates {(1,2) (2,2) (3,2) (4,2)};
    \addlegendentry{Evidence recoveries}
    \addplot[fill=vtaqua, draw=none] coordinates {(1,1) (2,1) (3,1) (4,1)};
    \addlegendentry{Corrections}
    \coordinate (evmid) at (axis cs:2.5,0);
  \end{axis}
  \node[subcap, anchor=north] at (evmid |- 0,-0.80) {(b) Harness events and release outcome};
\end{tikzpicture}
\caption{Verifier assertions and harness events of the four tool-use trials.}
\label{fig:app-trials}
\end{figure}
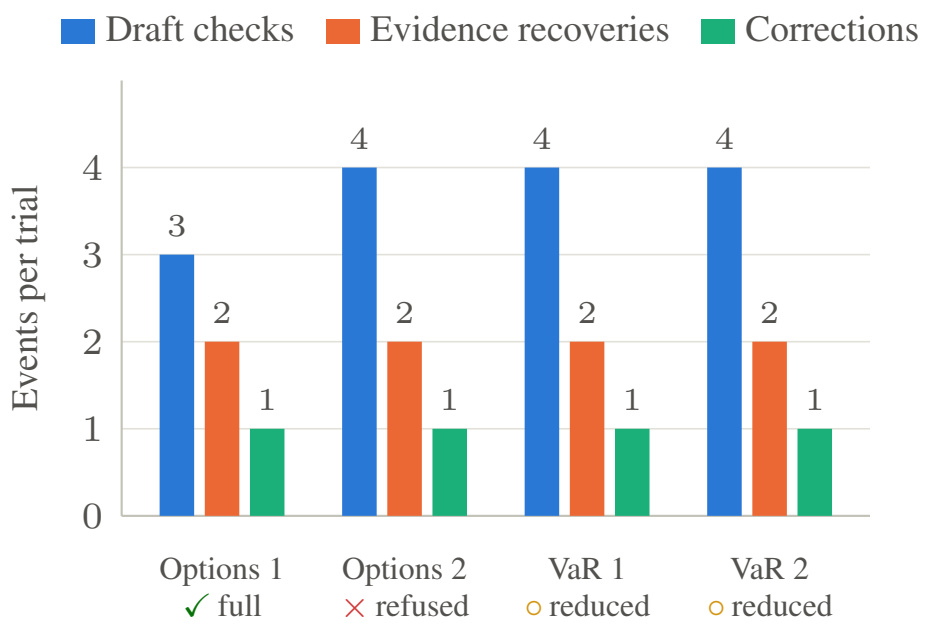

%% file: figures/eval/figF_claimvalues.tex
\begin{figure}[h]
\centering
\pgfplotsset{
  valid/.style={only marks, mark=*, mark size=2.3pt, vtblue, mark options={draw=white, line width=0.4pt}},
  wrong/.style={only marks, mark=diamond*, mark size=2.8pt, vtorange, mark options={draw=white, line width=0.4pt}},
  inv/.style={only marks, mark=triangle*, mark size=2.6pt, vtaqua, mark options={draw=white, line width=0.4pt}},
  span/.style={vtgrid, line width=1.6pt, forget plot},
  link/.style={densely dotted, vtmuted, line width=0.5pt, forget plot}}
\begin{tikzpicture}
\begin{groupplot}[vtaxis,
  group style={group size=4 by 1, horizontal sep=0.42cm, yticklabels at=edge left},
  scale only axis, width=2.62cm, height=1.25cm,
  ymin=0.45, ymax=2.55, ytick={1,2}, yticklabels={Equal weight, Risk parity},
  axis x line*=bottom, axis y line*=left, y axis line style={draw=none}, ytick style={draw=none},
  xmajorgrids, ymajorgrids, clip=false,
  xticklabel style={/pgf/number format/fixed, /pgf/number format/precision=2, /pgf/number format/fixed zerofill}]
\nextgroupplot[xmin=1.10, xmax=1.36, xtick={1.1,1.2,1.3}, xlabel={Sortino ratio},
  legend to name=claimleg, legend columns=3, legend style={/tikz/every even column/.append style={column sep=10pt}}]
  \addplot[span] coordinates {(1.1329,2) (1.2564,2)}; \addplot[span] coordinates {(1.1329,1) (1.3350,1)};
  \addplot[link] coordinates {(1.1329,2) (1.1329,1)}; \addplot[link] coordinates {(1.2115,2) (1.2115,1)};
  \addplot[valid] coordinates {(1.1329,2) (1.2115,1)}; \addlegendentry{Valid claim}
  \addplot[wrong] coordinates {(1.2115,2) (1.1329,1)}; \addlegendentry{Wrong-run claim (value of the other run)}
  \addplot[inv] coordinates {(1.2564,2) (1.3350,1)}; \addlegendentry{Invented claim}
\nextgroupplot[xmin=0.62, xmax=0.79, xtick={0.65,0.70,0.75}, xlabel={Calmar ratio}]
  \addplot[span] coordinates {(0.6422,2) (0.7657,2)}; \addplot[span] coordinates {(0.6422,1) (0.7687,1)};
  \addplot[link] coordinates {(0.6422,2) (0.6422,1)}; \addplot[link] coordinates {(0.6452,2) (0.6452,1)};
  \addplot[valid] coordinates {(0.6422,2) (0.6452,1)};
  \addplot[wrong] coordinates {(0.6452,2) (0.6422,1)};
  \addplot[inv] coordinates {(0.7657,2) (0.7687,1)};
\nextgroupplot[xmin=1.00, xmax=1.53, xtick={1.0,1.2,1.4}, xlabel={Turnover}]
  \addplot[span] coordinates {(1.0451,2) (1.4899,2)}; \addplot[span] coordinates {(1.0451,1) (1.3665,1)};
  \addplot[link] coordinates {(1.3665,2) (1.3665,1)}; \addplot[link] coordinates {(1.0451,2) (1.0451,1)};
  \addplot[valid] coordinates {(1.3665,2) (1.0451,1)};
  \addplot[wrong] coordinates {(1.0451,2) (1.3665,1)};
  \addplot[inv] coordinates {(1.4899,2) (1.1686,1)};
\nextgroupplot[xmin=0.68, xmax=0.84, xtick={0.70,0.75,0.80}, xlabel={Sharpe ratio}]
  \addplot[span] coordinates {(0.6936,2) (0.8170,2)}; \addplot[span] coordinates {(0.6936,1) (0.8278,1)};
  \addplot[link] coordinates {(0.6936,2) (0.6936,1)}; \addplot[link] coordinates {(0.7043,2) (0.7043,1)};
  \addplot[valid] coordinates {(0.6936,2) (0.7043,1)};
  \addplot[wrong] coordinates {(0.7043,2) (0.6936,1)};
  \addplot[inv] coordinates {(0.8170,2) (0.8278,1)};
\end{groupplot}
\node[anchor=north] at ($($(group c1r1.outer south)!0.5!(group c4r1.outer south)$)+(0,-2pt)$) {\pgfplotslegendfromname{claimleg}};
\end{tikzpicture}
\caption{Claimed values of the valid, wrong-run and invented claims in the evidence controls.}
\label{fig:app-claims}
\end{figure}
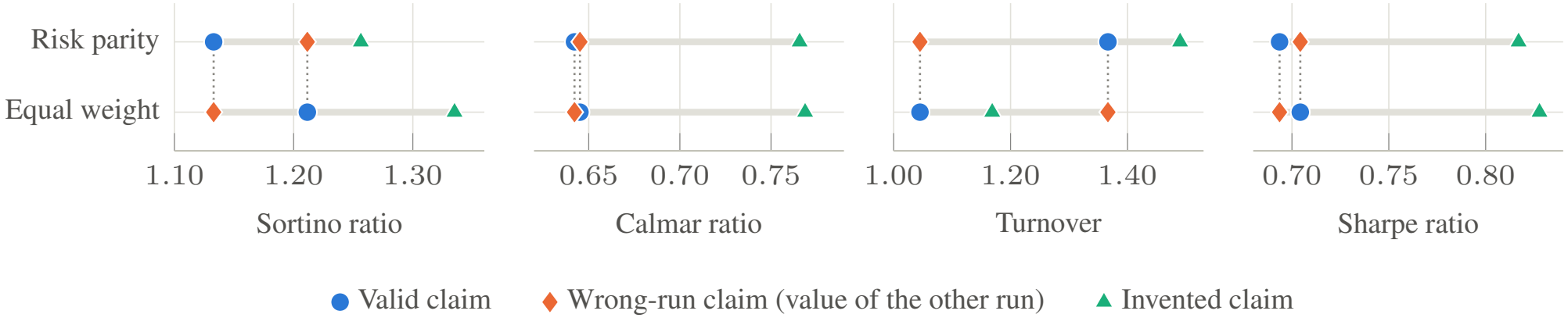